%% file: main.tex
\documentclass[10pt,twocolumn,letterpaper]{article}

\usepackage[pagenumbers]{cvpr}
\input{preamble}

\definecolor{cvprblue}{rgb}{0.21,0.49,0.74}
\usepackage[pagebackref,breaklinks,colorlinks,allcolors=cvprblue]{hyperref}

\def\paperID{35976} 
\def\confName{CVPR}
\def\confYear{2026}

\titleheader{\darkgrayed{This paper has been accepted for publication at the\\IEEE Conference on Computer Vision and Pattern Recognition (CVPR) Findings, Denver, 2026
\copyright IEEE}}
\title{Generative Event Pretraining with Foundation Model Alignment}

\author{Jianwen Cao
\and
Jiaxu Xing
\and
Nico Messikommer
\and
Davide Scaramuzza
\and \\
Robotics and Perception Group, University of Zurich
}

\begin{document}
\maketitle
\input{sec/0_abstract}

\input{sec/1_intro}
\input{sec/2_related}
\input{sec/3_method}

\input{sec/4_experiments}

\input{sec/6_conclusion}

\input{sec/5_ack}
\input{sec/7_supplementary}
\clearpage
{
    \small
    \bibliographystyle{ieeenat_fullname}
    \bibliography{main}
}
\end{document}

%% file: preamble.tex
\usepackage{amsmath,amssymb}
\usepackage{graphicx}
\usepackage{booktabs}
\usepackage{multirow}
\usepackage[table]{xcolor}
\usepackage{algorithmic}



%% file: sec/0_abstract.tex
\begin{abstract}
Event cameras provide robust visual signals under fast motion and challenging illumination thanks to their microsecond latency and high dynamic range.
However, their unique sensing characteristics and limited labeled data make it challenging to train event-based visual foundation models (VFMs), which are crucial for learning visual features transferable across tasks.
To tackle this problem, we propose GEP (Generative Event Pretraining), a two-stage framework that transfers semantic knowledge learned from internet-scale image datasets to event data while learning event-specific temporal dynamics.
First, an event encoder is aligned to a frozen VFM through a joint regression-contrastive objective, grounding event features in image semantics.
Second, a transformer backbone is autoregressively pretrained on mixed event–image sequences to capture the temporal structure unique to events.
Our approach outperforms state-of-the-art event pretraining methods on a diverse range of downstream tasks, including object recognition, segmentation, and depth estimation. 
Together, VFM-guided alignment and generative sequence modeling yield a semantically rich, temporally aware event model that generalizes robustly across domains.
Code: \href{https://github.com/uzh-rpg/generative_event_pretraining}{uzh-rpg/generative\_event\_pretraining}
\end{abstract}

%% file: sec/1_intro.tex
\section{Introduction}
\begin{figure}[t]
\centering
\includegraphics[width=0.48\textwidth]{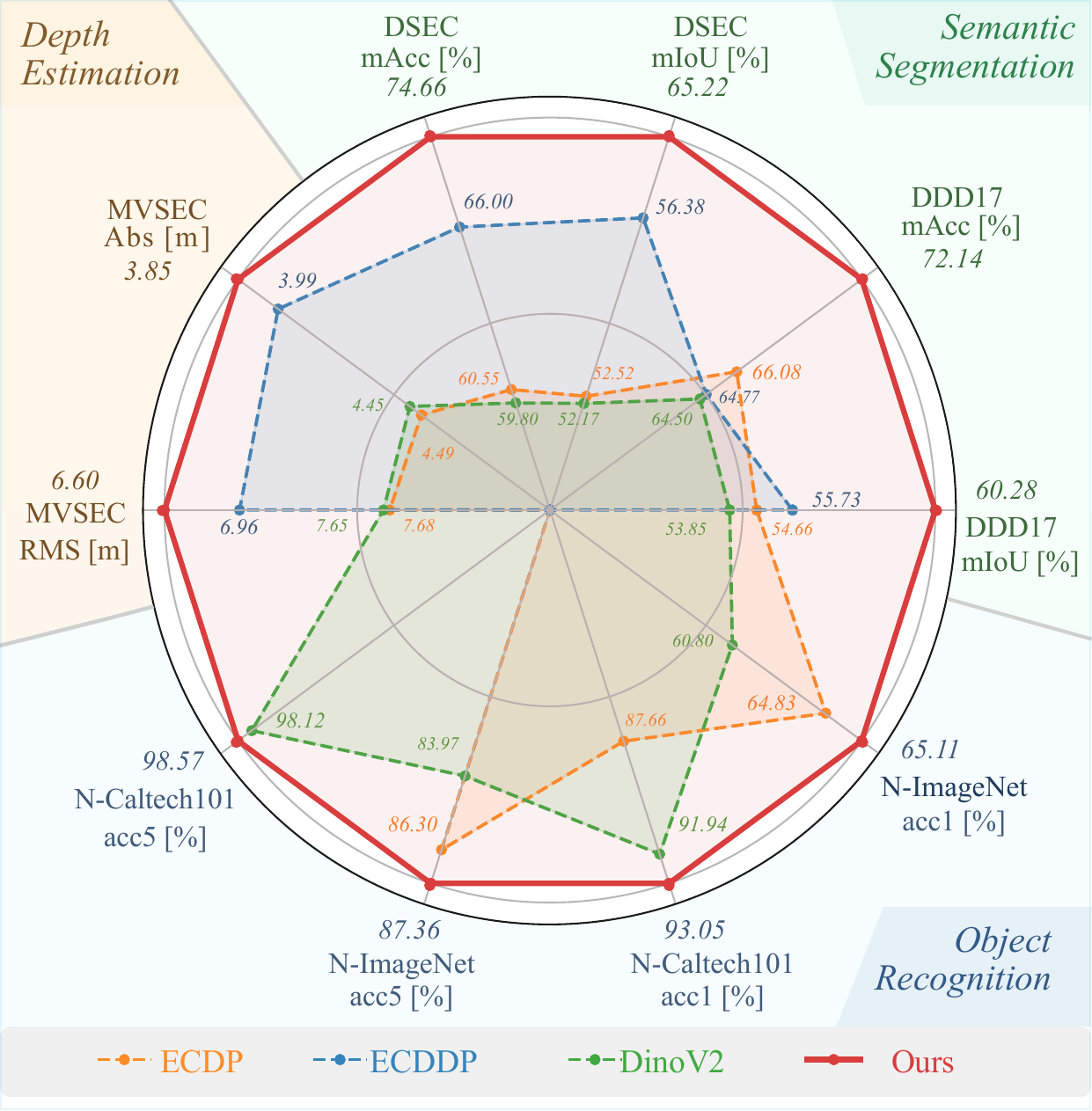}
\caption{Overall comparison across ten metrics on various datasets. With the same backbone model, our method demonstrates superior and consistent performance across all tasks.}
\label{fig:radar}
\end{figure}

Event cameras~\cite{lichtsteiner2008128} measure per-pixel brightness changes asynchronously with microsecond latency and high dynamic range.
Unlike conventional RGB frames, event streams are sparse and provide high temporal resolution, allowing robust perception in challenging lighting and fast-motion scenarios~\cite{gallego2020event, messikommer2026approximate}.
However, these advantages also introduce challenges: events contain limited texture, differ fundamentally from images, and lack access to large-scale training datasets~\cite{gallego2020event, rebecq2019high}. These challenges make it difficult to transfer semantic knowledge from images to events, and hinder the robustness of models that rely on large-scale, texture-rich supervision.

Consequently, many existing approaches rely on limited event-only pretraining, small task-specific datasets, or transfer directly from image-pretrained models that are not optimized for the sparse and asynchronous nature of event data~\cite{messikommer2022bridging, sun2022ess, yang2023event, yang2024event, klenk2024masked}. 
These strategies either miss long-range temporal dynamics or fail to import rich semantic priors from RGB due to suboptimal cross-modality adaptation.

While recent advances in vision foundation models (VFMs)~\cite{dosovitskiy2020image, radford2021learning, oquab2023dinov2, fang2023eva} demonstrate that large-scale pretraining on diverse image corpora produces semantically rich and task-transferable representations, these benefits have not yet been fully exploited in the event domain.
We address this gap with a unified pretraining paradigm that transfers VFM-level semantics to event data while explicitly modeling temporal dynamics in event streams.
By aligning event features with foundation-level image representations, the event encoder inherits broad semantic priors and narrows the discrepancy between event and image modalities.
The subsequent generative pretraining stage further equips the model with long-horizon predictive capabilities, enabling scalable, task-agnostic learning on large datasets featuring events and images.

To this end, we introduce \emph{Generative Event Pretraining (GEP)}, a two-stage training framework.
In the first stage, an event encoder is aligned to a frozen image encoder (DINOv2~\cite{oquab2023dinov2}) using a combination of MSE and InfoNCE~\cite{oord2018representation} objectives, transferring semantic structure from the image to the event domain. 
Such cross-modal alignment can also be viewed as a teacher-student
training paradigm~\cite{messikommer2025student}, where a pretrained teacher provides rich supervision for a student operating under
different observation modalities.
In the second stage, we perform autoregressive pretraining on aligned event and image sequences using a causal transformer. 
Unlike masked autoencoding~\cite{he2022masked}, our model predicts future token sequences over randomly masked temporal gaps, encouraging long-horizon temporal reasoning and providing dense generative supervision.
We pretrain on the large-scale Event-1.8M corpus with 1.8M event-image pairs, including EventScape~\cite{gehrig2021combining}, N-ImageNet~\cite{kim2021n}, and DSEC~\cite{gehrig2021dsec}, covering diverse spatial and temporal patterns. 

To evaluate generalization, we conduct experiments on object recognition (N-ImageNet~\cite{kim2021n}, N-Caltech101~\cite{orchard2015converting}), semantic segmentation (DDD17~\cite{binas2017ddd17}, DSEC~\cite{gehrig2021dsec}), and depth estimation (MVSEC~\cite{zhu2018multivehicle}). 
Among them, N-Caltech101, DDD17, and MVSEC are not included during pretraining and thus serve as out-of-distribution benchmarks. 

As summarized in Fig.~\ref{fig:radar}, our method consistently achieves superior performance across all 10 metrics, surpassing both event-specific and image-pretrained baselines while requiring only 24 pretraining epochs. 
These results demonstrate that aligning event and image features, coupled with autoregressive generative pretraining, enables effective cross-modal transfer and robust task generalization

The main contributions of our work are as follows:

\begin{enumerate}
    \item \emph{VFM-guided semantic alignment.}
    We align an event encoder with a pretrained VFM encoder to transfer semantic knowledge learned from large-scale image datasets, effectively grounding event features in rich visual priors.

    \item \emph{Autoregressive event pretraining.}
    We propose an autoregressive pretraining strategy on unlabeled event sequences paired with aligned images, which enables long-horizon temporal reasoning and captures the distinct temporal dynamics of event streams.

    \item \emph{Cross-domain generalization.} With about only 15\% training epochs compared to previous methods, our model outperforms state-of-the-art methods on recognition (N-ImageNet, N-Caltech101), segmentation (DDD17, DSEC), and depth estimation (MVSEC) benchmarks. Our pretrained backbone demonstrates robust generalization across diverse domains, providing a strong foundation for future event-based vision research.
\end{enumerate}

%% file: sec/2_related.tex
\section{Related Work}

\subsection{Transfer Learning Across Modalities}

In contrast to the conventional pretrain–finetune paradigm in the event domain, transferring knowledge across modalities involves aligning representations between images and events. Prior work in this direction can be broadly divided into two groups: Unsupervised domain adaptation and feature-level distillation.

\noindent \textbf{Unsupervised domain adaptation.}
UDA bridges the gap between labeled images and unlabeled events by enforcing consistency in features or predictions~\cite{messikommer2022bridging, sun2022ess, xie2024cross}. ESS~\cite{sun2022ess} aligns embeddings through reconstruction and prediction losses to transfer labels from Cityscapes to unpaired event data. CMESS~\cite{xie2024cross} adds attention-guided soft alignment and joint decoder constraints for dense prediction.

\noindent \textbf{Feature level distillation.}
Another line distills intermediate features from image models into event encoders using aligned image-event pairs. Hu et al.~\cite{hu2020learning} use grafting by replacing early image layers with an event front end and regressing internal features for label free adaptation. Depth AnyEvent~\cite{bartolomei2025depth} distills multi scale representations into event-based depth estimators and shows that semantic priors help reconstruction and estimation.

Across UDA and distillation, existing methods are designed for a single task and trained on limited data. In contrast, we align events to foundation-level image features in a task-agnostic way and couple the alignment with generative autoregressive pretraining, which captures temporal structure and enables transfer to both recognition and dense prediction tasks.

\subsection{Pretraining for Event Vision}
Event pretraining targets the learning of general feature representations that are subsequently finetuned for downstream tasks using task labels.

\noindent \textbf{Masked and self reconstruction pretraining.}
Masked Event Modeling adapts masked autoencoding and frame reconstruction to event streams. MEM~\cite{klenk2024masked} follows MAE~\cite{he2022masked} by masking patches and reconstructing from context. These methods capture local statistics but remain reconstruction-oriented and lack explicit temporal reasoning.

\noindent \textbf{Contrastive and multimodal alignment.}
Recent methods align events with images or with joint vision–language embedding spaces using contrastive objectives.
ECDP~\cite{yang2023event} aligns event and RGB patch embeddings through momentum-based contrastive learning, leveraging an image encoder pretrained with MoCov3~\cite{chen2021empirical} for supervision but without fully exploiting its rich semantic priors. ECDDP~\cite{yang2024event} extends this concept to spatial feature maps for pixel level tasks. EventCLIP~\cite{wu2023eventclip} projects event features into CLIP space for zero-shot and few-shot recognition. EventBind~\cite{zhou2024eventbind} applies multi-stage contrastive finetuning for stronger event image text alignment. In contrast, our framework adds feature regression toward foundation-level image representations and couples alignment with autoregressive pretraining, which yields semantically grounded and temporally aware event features.

\noindent \textbf{Autoregressive and predictive pretraining.}
Temporal modeling learns motion dynamics directly from events. Event Transformer~\cite{sabater2022event} introduces a sparse-aware transformer with patch-based representations and latent memory tokens for online recognition. Recent models replace recurrent blocks with structured state space modules S4 and S5~\cite{gu2021efficiently, zubic2024state}, which provide learnable time scales and adapt linear attention language modeling, such as RWKV~\cite{peng2024eagle}, for asynchronous encoding with multi-step or next representation prediction~\cite{hao2025maximizing}. These approaches confirm the strong temporal structure of events but are usually single modality, task-specific, and are often trained from scratch on limited data. We pretrain in a task-agnostic way under guidance from vision foundation models to unify semantic alignment with autoregressive temporal reasoning.

\subsection{Event VLMs with LLM backbones}
A recent popular direction toward generalizable event understanding is to extend large multimodal language models to event data for open-ended reasoning and description. EventGPT~\cite{liu2025eventgpt} aggregates spatiotemporal tokens before feeding a language model to support captioning and question answering. Event-VL~\cite{li2025eventvl} builds a generative event-based multimodal model with dynamic semantic alignment. EP VLM~\cite{qin2025event} uses event prior guided sparsification to prune redundant tokens and can adapt large backbones such as Qwen2-VL~\cite{wang2024qwen2}. Many event language and video language systems compress visual tokens by about an order of magnitude for efficiency, for example, EventGPT, Event-VL, VideoLLaMA3~\cite{zhang2025videollama}, and ARVideo~\cite{ren2024arvideo}. We observe that compression improves reasoning efficiency but harms dense prediction that requires fine spatial detail, see Sec.~\ref{sec:ablation}. Similar findings have been reported for segmentation and other dense tasks~\cite{tang2023dynamic, liu2024revisiting}. Large LLM backbones of billions of parameters also hinder real-time inference at high event rates. Our approach avoids heavy language backbones, preserves spatial detail, and transfers to recognition and dense prediction.

%% file: sec/3_method.tex
\section{Method}
\subsection{Overview}
\label{subsec:overview}
\begin{figure*}
\centering
\includegraphics[width=\textwidth]{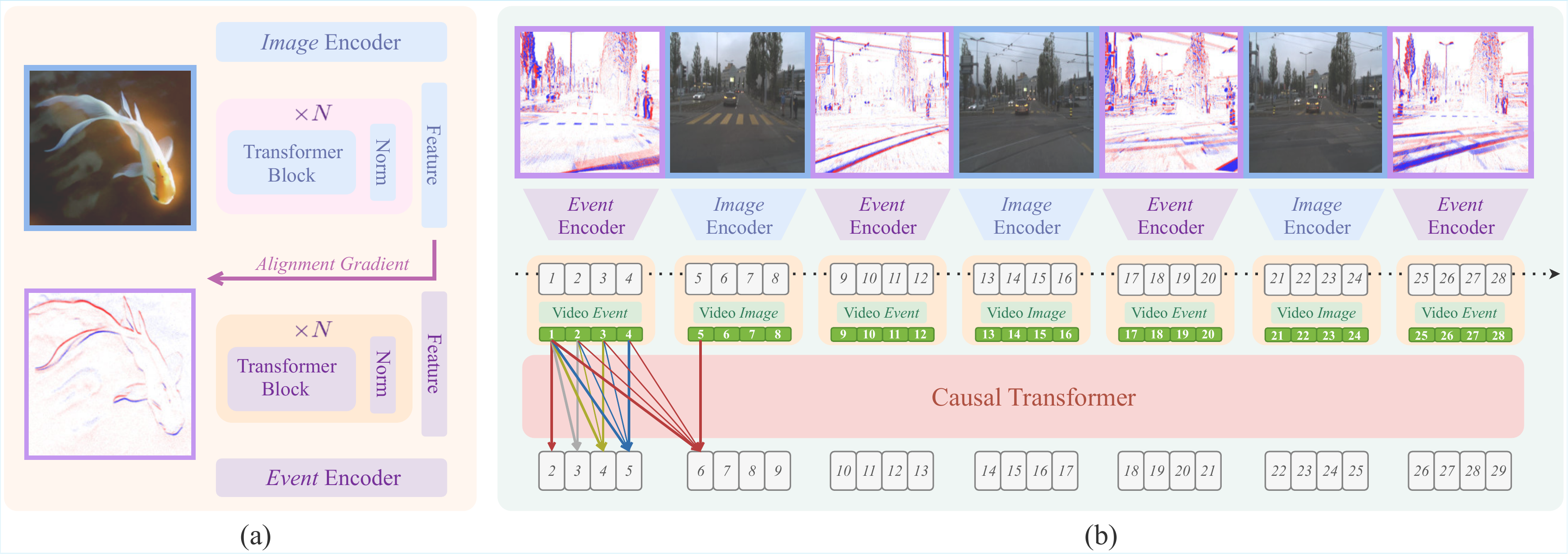}
\caption{The overall two-stage framework. 
(a) Alignment stage: Event frames and synchronized images are encoded by an event encoder and a frozen VFM encoder. 
The event encoder is optimized with regression, contrastive, and preservation terms to match the semantic structure of image features. 
(b) Autoregressive pretraining stage: Aligned event and image embeddings are interleaved into a single sequence and processed by a causal transformer, which predicts future slices from partial windows, learning long-range temporal structure and cross-modal consistency. Arrows indicate causal dependencies; only a subset is shown for visual clarity.}
\label{fig:framework}
\end{figure*}

Our method consists of two main stages: \emph{alignment} and \emph{pre-training}. 
An overview of the entire framework is illustrated in Fig.~\ref{fig:framework}.
Following the event accumulation as detailed in Sec~\ref{Event Accumulation}, each temporal window yields a normalized event frame $X_e$ paired with a synchronized image $X_i$, as illustrated in Fig.~\ref{fig:framework}\;(a). 
Given a paired event--image sample $(X_e, X_i)$, the event encoder $E_e$ and the image encoder $E_i$ extract modality-specific embeddings $(Z_e, Z_i)$ used for the modality alignment. 
For the pretraining, we alternately accumulate events and sample images over time, feeding them into the corresponding encoders to obtain a multimodal feature sequence.
All encoded embeddings are concatenated into a long sequence $S$, corresponding to the token sequence fed to the causal transformer in Fig.~\ref{fig:framework}\;(b). 
For each slice $S_{s, s+w}$, a causal transformer $T$ autoregressively predicts the next slice $S_{s+1, s+1+w}$:
\begin{equation}
    \hat S_{s+1, s+1+w}=T\big(S_{s, s+w}\big),
\end{equation}
where $s$ is a randomly sampled starting index and $w$ is a fixed window length.

\subsection{Event Accumulation}
\label{Event Accumulation}
Event streams are composed of asynchronous brightness changes represented as tuples $(x, y, t, p)$, where $(x, y)$ denotes pixel coordinates, $t$ the timestamp, and $p \in \{+1, -1\}$ the polarity indicating whether the brightness increased or decreased. 
To construct a dense tensor representation suitable for transformer-based modeling, we accumulate events within a fixed temporal window $\Delta t$ into a pseudo-frame $X_e \in \mathbb{R}^{H \times W \times 3}$.

For each event, we locate its corresponding pixel cell $(x, y)$ and increment the first channel count $M_r(x, y)$ if the polarity is positive ($p = +1$), or the third channel count $M_b(x, y)$ if it is negative ($p = -1$). 
For the second channel, we assign full intensity whenever any event occurs at $(x, y)$ within the temporal window, 
serving as an activity mask that marks all active pixels regardless of polarity.
After processing all events in the window, the accumulated counts are normalized to suppress local activity spikes and maintain a balanced dynamic range.

Let $M_c(x, y)$ denote the raw accumulated count in channel $c \in \{r, b\}$. We compute a normalization scale $\alpha_n$ as the $n$-th percentile of all pixel values across both channels (by default $n = 99$). Each channel is then clipped and normalized as
\begin{equation}
    X_e(x, y, c) = \frac{\min(M_c(x, y), \alpha_n)}{\alpha_n}, \quad c \in \{r, b\}.
\end{equation}
This percentile-based clipping and normalization ensure robust scaling and prevent a few high-activity pixels from dominating the distribution, resulting in a stable and balanced input representation for subsequent encoding.

At the end of this stage, each event window produces a normalized event frame $X_e$ that can be paired with a synchronized image frame $X_i$. These paired samples $(X_e, X_i)$ serve as inputs to the following alignment and pretraining stages.

\subsection{Alignment}
The weights of both encoders are initialized from \mbox{DINOv2~\cite{oquab2023dinov2}}. 
We freeze the image branch $E_i$ (parameters $\theta_i$) and update only the event branch $E_e$ (parameters $\theta_e$). 
Since vision foundation models already capture strong and well-structured semantics from large-scale RGB corpora, 
their features provide a reliable reference. 
Freezing $E_i$ therefore treats the image encoder as a fixed semantic teacher, allowing the event encoder to learn meaningful alignment instead of co-adapting to unstable event features. 
The alignment objective combines a cosine similarity and an InfoNCE term, while a preservation loss constrains the behavior of $E_e$ on image inputs. 
Unless otherwise noted, we use DINOv2~\cite{oquab2023dinov2} as the image foundation model.

\noindent \textbf{Event--image alignment.}
Given a paired input $(X_e, X_i)$, we extract features
\[
Z_e = E_e(X_e), \qquad Z_i = \operatorname{sg}\!\big(E_i(X_i)\big),
\]
where $\operatorname{sg}$ stops gradients through the frozen image branch. 
We use the following alignment loss
\begin{equation}
\label{eq:align}
\mathcal{L}_{\text{a}}
= 
\lambda_{\text{cos}} \,\big(1 - \mathrm{cos}(Z_e, Z_i)\big)
+ 
\lambda_{\text{nce}} \,\mathcal{L}_{\text{nce}}\!\left(Z_e, Z_i\right),
\end{equation}
where $\mathrm{cos}(\cdot,\cdot)$ denotes cosine similarity. Following standard in-batch contrastive learning, the InfoNCE loss is defined as
\begin{equation}
\label{eq:nce}
\mathcal{L}_{\text{nce}}
= -\frac{1}{N} \sum_{n=1}^{N}
\log
\frac{\exp\!\big(\mathrm{cos}(\tilde z_{e}^{\,n}, \tilde z_{i}^{\,n}) / \tau \big)}
{\sum_{m=1}^{N} \exp\!\big(\mathrm{cos}(\tilde z_{e}^{\,n}, \tilde z_{i}^{\,m}) / \tau \big)},
\end{equation}
where $\tilde z = z / \lVert z \rVert_2$ denotes $\ell_2$-normalized features, $\tau$ is the temperature, 
and $N$ is the batch size. Each positive pair $(\tilde z_e^{\,n}, \tilde z_i^{\,n})$ is contrasted against all other image embeddings in the batch, which act as negative anchors.

The cosine term enforces directional consistency between event and image embeddings, ensuring that both modalities lie in a shared feature subspace regardless of scale, 
while the InfoNCE term encourages global separation from unrelated image embeddings, leading to semantically consistent yet discriminative representations. 
In practice, we apply a lightweight projection head $g(\cdot)$ before \eqref{eq:nce} following SimCLR~\cite{chen2020simple}. 
This allows the contrastive branch to adapt its embedding distribution without disturbing the main cosine alignment space, stabilizing optimization and improving cross-modal transfer.

\noindent \textbf{Capability preservation on images.}
To avoid drift when $E_e$ processes RGB inputs, we pass $X_i$ through $E_e$ and match the frozen image features
\[
Z_e^{(I)} = E_e(X_i), \qquad Z_i = \operatorname{sg}\!\big(E_i(X_i)\big),
\]
with the preservation loss
\begin{equation}
\label{eq:pres}
\mathcal{L}_{\text{p}} = \mu \,\big(1 - \mathrm{cos}(Z_e^{(I)}, Z_i)\big).
\end{equation}
Equation~\eqref{eq:pres} regularizes $E_e$ by preventing its feature space from collapsing or drifting away from the semantic manifold learned by $E_i$. 
By enforcing consistent orientation when both branches process image inputs, this term discourages degenerate alignment and maintains semantic grounding.

\noindent \textbf{Total objective.}
The alignment stage optimizes
\begin{equation}
\label{eq:total}
\mathcal{L} = \mathcal{L}_{\text{a}} + \mathcal{L}_{\text{p}},
\qquad
\lambda_{\text{cos}} > 0,\; \lambda_{\text{nce}} > 0,\; \mu > 0.
\end{equation}
To balance the loss terms, we use separate weights for cosine alignment ($\lambda_{\text{cos}}$), contrastive separation ($\lambda_{\text{nce}}$), and preservation ($\mu$). 
Batch composition and temperature $\tau$ follow standard practice for in-batch contrastive learning.

\subsection{Pre-training}
After alignment, both encoders produce semantically aligned feature embeddings.
We concatenate them into a multimodal sequence $S = [S_1, S_2, \dots, S_K]$
that mixes samples from image datasets and paired event-video datasets, as depicted in Fig.~\ref{fig:framework}\;(b).
Each token is projected into a shared embedding space and augmented with
positional and modality encodings before entering the causal transformer.

\noindent \textbf{Token representation.}
Given an encoded token $S_k$, its input representation is
\begin{equation}
    X_k = S_k + P_k + M_k ,
\end{equation}
where $P_k$ is the positional embedding and $M_k$ is the modality encoding,
indicating whether the token originates from an event frame, image frame, or video sequence.
The modality encoding allows the transformer to distinguish heterogeneous sources within the same batch, while positional encoding preserves temporal order.

\noindent \textbf{Dense autoregressive training.}
We slice a long multimodal sequence into overlapping windows of length $w$ with a stride $1$ to form a dense training target.
In other words, within one training sample, the model performs
\[
S_1 \!\to\! S_2, \quad
(S_1,S_2) \!\to\! S_3, \quad \dots, \quad
S_{1:w-1} \!\to\! S_w ,
\]
yielding a dense sequence of autoregressive targets:
\begin{equation}
\mathcal{L}_{\text{pre}}
= \frac{1}{w}\sum_{j=1}^{w}
\big\|
\hat{S}_{s+j} - S_{s+1+j}
\big\|_2^2 ,
\end{equation}
encouraging temporally coherent and modality-consistent predictions.

%% file: sec/4_experiments.tex
\section{Experiments}
\subsection{Datasets and Baselines}
\label{sec:datasets}
We evaluate on established event-vision benchmarks that cover both recognition, segmentation, and depth estimation tasks. 
For object recognition, we use N-ImageNet~\cite{kim2021n} and N-Caltech101~\cite{orchard2015converting}, the neuromorphic counterparts of ImageNet~\cite{deng2009imagenet} and Caltech101~\cite{fei2004learning}. 
N-ImageNet contains large-scale event recordings of 1K ImageNet categories captured with a moving DAVIS sensor~\cite{brandli2014240}, while N-Caltech101 provides 101 object classes with simpler backgrounds and fewer samples per class. 
For semantic segmentation, we use DDD17~\cite{binas2017ddd17} and DSEC~\cite{gehrig2021dsec, sun2022ess}, two large-scale driving datasets with synchronized events and grayscale frames. 
DDD17 offers 12 hours of urban driving sequences with 6 semantic classes derived from intensity frames, while DSEC provides 8,082 higher-resolution stereo event-image-annotation pairs for training.
For depth estimation, we use the MVSEC dataset~\cite{zhu2018multivehicle}, which provides synchronized stereo DAVIS346 event cameras with grayscale frames and accurate depth maps.
MVSEC contains indoor and outdoor sequences captured from handheld, hexacopter, car, and motorcycle platforms, enabling depth evaluation under diverse motions and illumination conditions.

For pretraining, we adopt Event-1.8M, a large-scale event corpus that integrates EventScape~\cite{gehrig2021combining}, N-ImageNet, and DSEC. 
This mixture covers diverse spatial patterns and motion statistics, serving as a task-agnostic dataset for representation learning. 
DDD17, MVSEC, and N-Caltech101 are excluded from pretraining and used only for out-of-distribution evaluation. 
More training details are included in the supplementary material.

We compare our model against four groups of baselines to ensure a fair and comprehensive evaluation. 
The first group trains a ViT~\cite{dosovitskiy2020image} from scratch on event data without any pretraining.
The second group is specifically designed for each task.
The third group includes self-supervised pretraining on images, represented by MAE~\cite{he2022masked}, BeiT~\cite{bao2021beit}, and DINOv2~\cite{oquab2023dinov2}. 
The last group consists of event-specific pretraining methods, including MEM~\cite{klenk2024masked}, ECDP~\cite{yang2023event}, ECDDP~\cite{yang2024event}, and EventBind~\cite{zhou2024eventbind}.
\subsection{Object Recognition}
\begin{table}
    \centering
    \caption{Object recognition on N-ImageNet and N-Caltech101. We report top-1 (acc1) and top-5 (acc5) accuracies. The two best-performing methods for each evaluation metric are highlighted in \colorbox{YellowGreen!25}{green} and \colorbox{Orange!10}{orange}.}
    \label{tab:object_recognition_results}
    \begingroup
    \setlength{\tabcolsep}{2pt}
    \resizebox{\linewidth}{!}{%
    \begin{tabular}{lllccccc}
        \toprule
        Method & Backbone & Dataset & Ep. & 
        \multicolumn{2}{c}{N-ImageNet} & 
        \multicolumn{2}{c}{N-Caltech101} \\
        & & & & acc1$\uparrow$ & acc5$\uparrow$ & acc1$\uparrow$ & acc5$\uparrow$ \\
        \midrule
        \rowcolor{cvprblue!8}\multicolumn{8}{l}{\textit{Training from scratch}} \\
        ViT\cite{dosovitskiy2020image}  & ViT-S/16 & N-ImageNet  & 300 & 46.70 & 69.89 & 55.63 & -- \\
        \rowcolor{cvprblue!8}\multicolumn{8}{l}{\textit{Specific trained}} \\
        EST\cite{gehrig2019end}  & -- & -- & -- & 48.93 & -- & -- & -- \\
        \rowcolor{cvprblue!8}\multicolumn{8}{l}{\textit{Self-supervised pretraining}} \\
        BeiT\cite{bao2021beit} & ViT-B/16 & ImageNet-1K & 800 & 47.15 & 69.27 & 53.10 & -- \\
        MAE\cite{he2022masked}  & ViT-B/16 & ImageNet-1K & 800 & 51.25 & 72.64 & 67.68 & -- \\
        MoCo-v3\cite{chen2021empirical}  & ViT-S/16 & ImageNet-1K & 300 & 45.77 & 68.89 & 76.59 & -- \\
        DINOv2\cite{oquab2023dinov2}  & ViT-S/16 & LVD-142M & -- & 60.80 & 83.97 & 91.94 & 98.12 \\
       \rowcolor{cvprblue!8}\multicolumn{8}{l}{\textit{Event-specific pretraining}} \\
        MEM\cite{klenk2024masked}  & ViT-S/16 & N-ImageNet & 75 & 57.89 & -- & -- & -- \\
        ECDP\cite{yang2023event} & ViT-S/16 & N-ImageNet & 300 & 64.83 & 86.30 & 87.66 & -- \\
        EventBind\cite{yang2024event} & ViT-B/16 & N-ImageNet & -- & 51.40 & -- & \cellcolor{Orange!10}94.08 & -- \\
        STP~\cite{liang2025efficient} & Swin-T & N-ImageNet & -- & 68.87 & 89.65 & -- & -- \\
        \midrule
        Ours & ViT-S/16 & Event-1.8M & \cellcolor{Orange!10}24 
            & 65.11 & 87.36 & 93.05 & \cellcolor{Orange!10}98.57 \\
        Ours (Sup. Align) & ViT-S/16 & N-ImageNet & \cellcolor{YellowGreen!25}6 
            & \cellcolor{Orange!10}69.41 & \cellcolor{Orange!10}90.20 & -- & -- \\
        Ours& ViT-B/16 & Event-1.8M & \cellcolor{Orange!10}24 
            & \cellcolor{YellowGreen!25}75.20 & \cellcolor{YellowGreen!25}92.90 & \cellcolor{YellowGreen!25}96.47 & \cellcolor{YellowGreen!25}99.56 \\
        \bottomrule
    \end{tabular}%
    }
    \endgroup
\end{table}
As can be observed by the top-1 and top-5 accuracy reported in Table~\ref{tab:object_recognition_results}, our model consistently surpasses event-specific pretraining baselines. 
Compared with ECDP, we improve top-1 from 64.83\% to 65.11\% on N-ImageNet and from 87.66\% to 93.05\% on N-Caltech101. 
Although the event encoder is aligned to DINOv2, our model achieves higher accuracy on event streams than its VFM teacher, trained on 142M RGB images and applied to events.
This shows the benefit of the alignment followed by autoregressive modeling to transfer semantics while adapting the features to the sparsity and asynchronicity of events. 
With only 24 epochs of training (including alignment), our approach outperforms \mbox{DINOv2} on the tested event benchmarks, demonstrating strong data efficiency and generalization.
Remarkably, our method outperforms STP~\cite{liang2025efficient} under fair settings, while using only 25\% of the pre-training schedule (Ours Sup. Align). 
As shown in Table~\ref{tab:object_recognition_results}, STP's parameter efficiency relies on privileged supervised pre-training with explicit labels.

\begin{table}[t]
    \centering
    \caption{Semantic segmentation on DDD17 and DSEC. We report mean IoU (mIoU) and mean ACC (mAcc). The two best-performing methods for each evaluation metric are highlighted in green and orange.}
    \label{tab:semantic_segmentation_results}
    \begingroup
    \footnotesize
    \setlength{\tabcolsep}{2pt}
    \resizebox{\linewidth}{!}{%
    \begin{tabular}{lllcccccc}
        \toprule
        {Method} & {Backbone} & {Dataset} & {Ep.} &
        \multicolumn{2}{c}{{DDD17}} & \multicolumn{2}{c}{{DSEC}} \\
        & & & & {mIoU$\uparrow$} & {mAcc$\uparrow$} & {mIoU$\uparrow$} & {mAcc$\uparrow$} \\
        \midrule
        \rowcolor{cvprblue!8}\multicolumn{8}{l}{\textit{Training from scratch}} \\
        ViT\cite{dosovitskiy2020image}   & ViT-S/16 & -- & -- & 36.65 & 46.21 & 32.66 & 40.84 \\
        \rowcolor{cvprblue!8}
        \multicolumn{8}{l}{\textit{Specific trained}} \\
        ESS\cite{sun2022ess}   & -- & Cityscape & 50 & \cellcolor{Orange!10}61.37 & 70.87 & 53.30 & 62.94 \\
        \rowcolor{cvprblue!8}
        \multicolumn{8}{l}{\textit{Self-supervised pretraining}} \\
        BeiT\cite{bao2021beit}  & ViT-B/16 & ImageNet1K & 800 & 52.39 & 61.95 & 51.90 & 59.66 \\
        MAE\cite{he2022masked}   & ViT-B/16 & ImageNet1K & 800 & 53.76 & 64.78 & 51.96 & 59.84 \\
        DINOv2\cite{oquab2023dinov2}& ViT-S/14 & LVD-142M & -- & 53.85 & 64.50 & 52.17 & 59.80 \\
        \rowcolor{cvprblue!8}
        \multicolumn{8}{l}{\textit{Event-specific pretraining}} \\
        MEM\cite{klenk2024masked}   & ViT-S/16 & N-ImageNet & 75 & -- & -- & 44.62 & 51.39 \\
        ECDP\cite{yang2023event}  & ViT-S/16 & N-ImageNet & 300 & 54.66 & 66.08 & 52.52 & 60.55 \\
        ECDDP\cite{yang2024event} & ViT-S/16 & E-TartanAir & 300 & 55.73 & 64.77 & 56.38 & 66.00 \\
        TESPEC~\cite{mohammadi2025tespec} & Swin-T & 1Mpx~\cite{perot2020learning} & -- & -- & -- & 62.77 & 70.61 \\
        \midrule
        Ours & {ViT-S/14} & {N-ImageNet} & \cellcolor{YellowGreen!25}{24}
            & 58.42 & 69.59 & {60.43} & {68.76} \\
        Ours & {ViT-S/14} & {Event-1.8M} &\cellcolor{YellowGreen!25}{24}
            & {60.28} &\cellcolor{Orange!10}{72.14} & \cellcolor{Orange!10}{65.22} & \cellcolor{Orange!10}{74.66} \\
        Ours & {ViT-B/14} & {Event-1.8M} &\cellcolor{YellowGreen!25}{24}
            & \cellcolor{YellowGreen!25}{61.90} & \cellcolor{YellowGreen!25}{72.39} &\cellcolor{YellowGreen!25}{67.37} &\cellcolor{YellowGreen!25}{76.27} \\
        \bottomrule
    \end{tabular}%
    }
    \endgroup
\end{table}

\subsection{Semantic Segmentation}
For DDD17, we use sequence1 for training and the remaining sequences for evaluation, as done in ESS~\cite{sun2022ess}.
Following DINOv2, a simple linear segmentation head is attached on top of the encoder.
Thus, the performance mainly reflect the quality of the backbone representations. We use cross-entropy and Dice losses~\cite{milletari2016v} on labeled events.

Note that both ECDP and ECDDP adopt a stronger UperNet~\cite{xiao2018unified} decoder.
ECDP further incorporates a 3D expanded patch embedding~\cite{yue2021vision},
while ECDDP additionally applies test-time augmentation with horizontal flipping and multi-scale features.
Despite our lighter linear head and the absence of test time augmentation, our method achieves higher segmentation accuracy, indicating that the gains come from the learned representations rather than from a heavy decoder or inference heuristics.

\begin{table}[t]
    \centering
    \caption{Depth estimation on MVSEC. We report absolute error (Abs↓) and root mean squared error (RMS↓). The two best-performing methods for each evaluation metric are highlighted in green and orange.}
    \label{tab:depth_estimation_results}
    \begingroup
    \footnotesize
    \setlength{\tabcolsep}{3pt}
    \renewcommand{\arraystretch}{1.05}
    \resizebox{0.9\linewidth}{!}{%
    \begin{tabular}{lllccc}
        \toprule
        Method & Backbone & Dataset & Ep. & Abs$\downarrow$ & RMS$\downarrow$ \\
        \midrule
        \rowcolor{cvprblue!8}
        \multicolumn{6}{l}{\textit{Specific trained}} \\
        HMNet\cite{hamaguchi2023hierarchical} & -- & -- & -- & 4.61 & 8.60 \\
        \rowcolor{cvprblue!8}
        \multicolumn{6}{l}{\textit{Self-supervised pretraining}} \\
        BeiT\cite{bao2021beit} & ViT-B/16 & ImageNet-1K & 800 & 4.40 & 7.56 \\
        MAE\cite{he2022masked} & ViT-B/16 & ImageNet-1K & 800 & 4.45 & 7.60 \\
        DINOv2\cite{oquab2023dinov2} & ViT-S/16 & LVD-142M & -- & 4.45 & 7.65 \\
        \rowcolor{cvprblue!8}
        \multicolumn{6}{l}{\textit{Event-specific pretraining}} \\
        ECDP\cite{yang2023event} & ViT-S/16 & N-ImageNet & 300 & 4.49 & 7.68 \\
        ECDDP\cite{yang2024event} & ViT-S/16 & N-ImageNet & 300 & \cellcolor{Orange!10}{3.99} & \cellcolor{Orange!10}{6.96} \\
        \midrule
        Ours & ViT-S/16 & Event-1.8M & \cellcolor{YellowGreen!25}{24} & \cellcolor{YellowGreen!25}{3.85} & \cellcolor{YellowGreen!25}{6.60} \\
        \bottomrule
    \end{tabular}%
    }
    \endgroup
\end{table}

Table~\ref{tab:semantic_segmentation_results} summarizes the semantic segmentation performance reported in mIoU and mAcc. 
Our model attains the best performance on both datasets, outperforming ECDDP by +4.0 mIoU and +5.0 mAcc on DSEC, indicating that purely static pretraining struggles to capture event dynamics, while our alignment plus autoregressive pretraining bridges this gap.

\begin{figure}[b]
    \centering
    \includegraphics[width=0.48\textwidth]{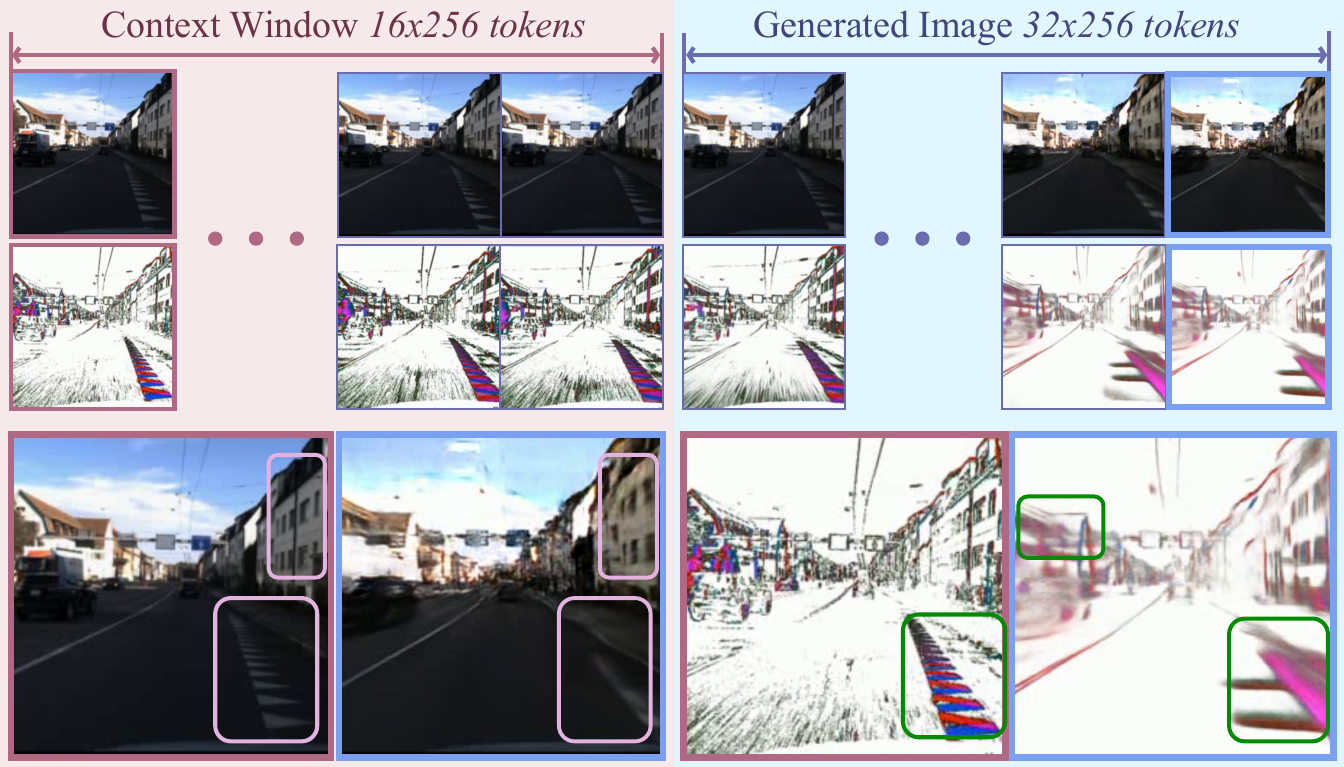}
    \caption{Visualization of a 16-frame context (blue) and a 32-frame autoregressively generated future (red) on validation interleaved event and image streams. Green boxes highlight consistent motion. Insets enlarge the first context and last generated frames for clarity.}
    \label{fig:pretrain_vis}
\end{figure}

\begin{figure*}[ht]
    \centering
    \includegraphics[width=0.98\textwidth]{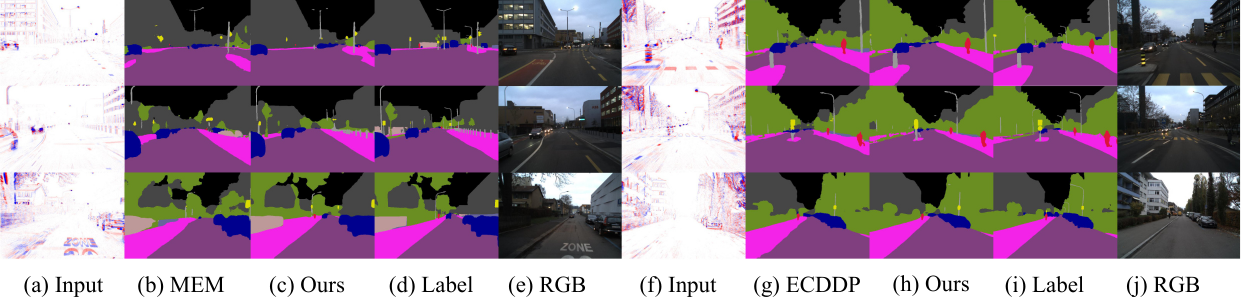}
    \caption{Qualitative results on the DSEC validation set. The RGB images are shown only for visual reference and are not used by the model.}
    \label{fig:dsec_qualitative}
\end{figure*}
\subsection{Depth estimation}
We follow the protocol of ECDDP~\cite{yang2024event} on MVSEC.
Training is conducted on the 'outdoor\_day2' split, and validation is performed on 'outdoor\_day1', 'outdoor\_night1', 'outdoor\_night2', and 'outdoor\_night3'.
Following ECDDP, the network is trained to predict normalized log depth.
The loss combines a scale-invariant term and a multi-scale gradient matching term.
At test time, predictions are converted back to metric depth, and we report Absolute error and Root Mean Squared error in meters without limiting the maximum depth during evaluation, consistent with prior work~\cite{gehrig2021combining}.
\begin{figure}[b]
    \centering
    \includegraphics[width=0.47\textwidth]{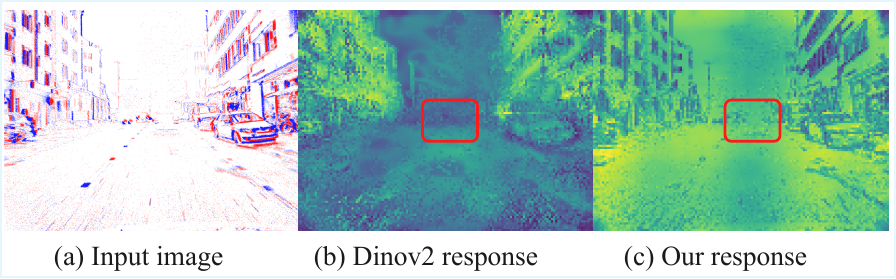}
    \caption{Attention map response on event camera data.}
    \label{fig:attention}
\end{figure}
Table~\ref{tab:depth_estimation_results} summarizes the results.
Our method achieves lower Abs and RMS errors than ECDDP and other baselines, demonstrating that the proposed pretraining remains highly effective for depth estimation tasks that require strong 3D spatial understanding.

\subsection{Qualitative results}
\noindent \textbf{Autoregressive long-horizon generation.}
We visualize sequence rollouts using a frozen, lightweight decoder attached to the encoder, which is used only for qualitative inspection of latent predictions.  
Each rollout is conditioned on a multimodal context of $256{\times}8{\times}2=4096$ tokens and generates $256{\times}16{\times}2=8192$ future tokens.  
Here, 256 denotes spatial tokens per frame, 8 and 16 are the context and predicted steps, and 2 corresponds to interleaved event and image modalities.  
A sliding causal window is applied when the target horizon exceeds the pretraining window length.  
Figure~\ref{fig:pretrain_vis} indicates that the model extrapolates ego motion, facade parallax, lane-marking drift, and object displacement even when such cues are weak in the last context frame.  
Despite moderate sensor noise and the minimal decoder, rollouts remain stable and geometrically consistent, which suggests the transformer internalizes long-range temporal structure.

\noindent \textbf{Segmentation results.}
Figure~\ref{fig:dsec_qualitative} compares our model with MEM~\cite{klenk2024masked} and ECDDP\cite{yang2024event}.  
Our predictions exhibit cleaner boundaries and more coherent region semantics, especially in low-texture or overexposed scenes where event measurements are sparse.  
In the last row, the roadside pedestrian is barely discernible in the event input due to missing texture, which makes fine structures difficult to detect.  
With VFM-guided alignment and generative pretraining, the network recovers semantically consistent contours and preserves layout in spite of weak appearance cues.

We refer to the supplementary material for more qualitative task results

\noindent \textbf{Attention responses.}
Figure~\ref{fig:attention} analyzes attention on event inputs.  
The red box marks a distant car that is hard to perceive in the raw events.  
DINOv2 with registers~\cite{darcet2023vision} usually produces meaningful attention on RGB data, yet on events, its response is weak and spatially inconsistent with spurious activations on the road surface.  
Our aligned and generatively pretrained encoder yields concentrated and structured attention that focuses on the car and other salient elements, indicating effective cross-modal alignment and stronger semantic grounding.
\begin{figure}[b]
  \centering
  \includegraphics[width=0.48\textwidth]{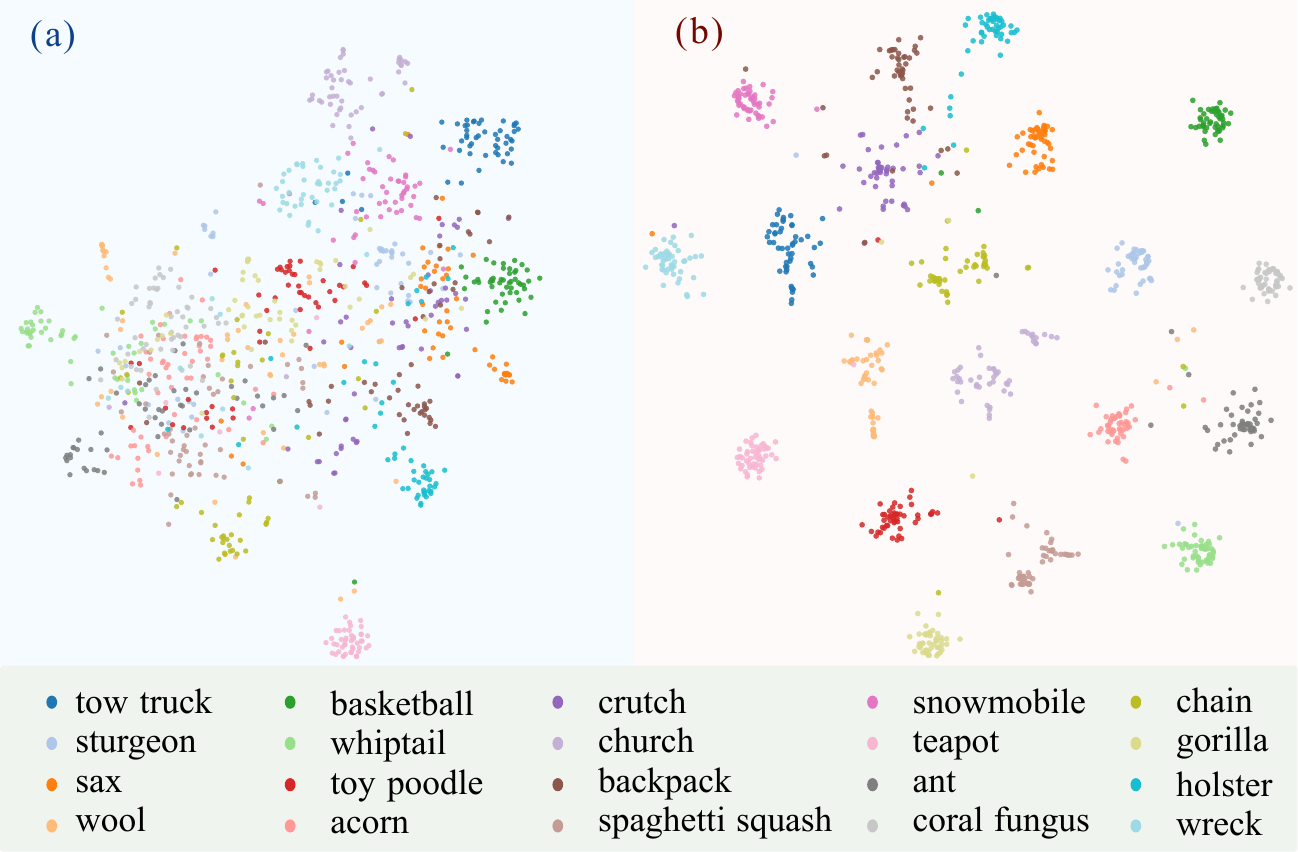}
  \caption{
  t-SNE visualization of ViT-Base encoder features 
  (20 classes, 50 samples/class from N-ImageNet validation set). 
  Left (a): \textit{Before Align}; right (b): \textit{After Align}. 
  Each color denotes a semantic class.
  }
  \label{fig:tsne}
\end{figure}

\begin{table}[t]
  \centering
  \caption{Clustering metrics on encoder features 
  (20 classes, 50 samples/class). 
  Higher is better for Silhouette and Calinski--Harabasz; 
  lower is better for Davies--Bouldin.}
  \label{tab:tsne_metrics}
  \resizebox{\linewidth}{!}{%
  \begin{tabular}{lcc|cc}
    \toprule
    & \multicolumn{2}{c}{ViT-Small} & \multicolumn{2}{c}{ViT-Base} \\
    \cmidrule(lr){2-3}\cmidrule(lr){4-5}
    Metric & Dinov2 & Aligned & Dinov2 & Aligned \\
    \midrule
    \rowcolor{gray!10}
    Silhouette\cite{rousseeuw1987silhouettes} $\uparrow$        & -0.03 & \textbf{0.19}   & 0.03  & \textbf{0.27} \\
    Davies--Bouldin\cite{davies2009cluster} $\downarrow$ & 5.24  & \textbf{2.21}   & 3.75  & \textbf{2.04}   \\
    \rowcolor{gray!10}
    Calinski--Harabasz\cite{calinski1974dendrite} $\uparrow$& 9.39  & \textbf{33.21}  & 14.97 & \textbf{48.56}  \\
    \bottomrule
  \end{tabular}%
  }
  \vspace{-0.4cm}
\end{table}

\noindent \textbf{Alignment Clustering.}
Figure~\ref{fig:tsne} shows a t-SNE plot for event features before and after alignment.
Although the alignment pretraining is class-agnostic, it still exhibits pronounced class-wise separation, consistent with near-linear separability in the learned feature space.
Table~\ref{tab:tsne_metrics} quantifies the effect, which indicates reduced intra-class variance and stronger inter-class separation, supporting the view that alignment organizes event features into semantically meaningful manifolds that aid downstream tasks.

\subsection{Ablations}
\label{sec:ablation}
We report linear probing with a frozen backbone to isolate representation quality in order to ablate alignment, the second-stage autoregressive modeling, masked reconstruction, and token aggregation.

\begin{table}[t]
  \centering
  \caption{Feature alignment results on N-ImageNet using 30,000 training steps. Each cell shows Original→Aligned with relative change (\%).}
  \label{tab:ablation_alignment_half}
  \begingroup
  \setlength{\tabcolsep}{4pt}
  \renewcommand{\arraystretch}{1.05}
  \footnotesize
  \resizebox{\linewidth}{!}{
  \begin{tabular}{lcc}
    \toprule
    \textbf{Metric} & \textbf{Distance Reduction} & \textbf{ImageNet Acc. (\%)} \\
    \midrule
    \rowcolor{gray!10}
    MSE              & 2.22→0.78 (-64.7\%) & 31.18→61.45 \\
    InfoNCE          & 2.28→0.57 (-75.0\%) & 31.18→60.99 \\
    \rowcolor{gray!10}
    Cosine           & 0.70→0.27 (-61.8\%) & 31.18→61.80 \\
    KL Diver.        & 0.48→0.15 (-67.5\%) & 31.18→59.20 \\
    \rowcolor{gray!10}
    Attention        & 38.76→19.42 (-49.9\%) & 31.18→56.72 \\
    Cosine + InfoNCE & 1.40→0.36 (-74.3\%) & 31.18→\textbf{62.73} \\
    \rowcolor{gray!10}
    All              & 0.96→0.37 (-61.1\%) & 31.18→61.40 \\ 
    \bottomrule
  \end{tabular}
  }
  \endgroup
\vspace{-0.4cm}
\end{table}

\noindent \textbf{Feature alignment analysis.}
Table~\ref{tab:ablation_alignment_half} systematically evaluates different alignment objectives on N-ImageNet.
Among the individual objectives, \emph{Cosine + InfoNCE} achieves the best trade-off between compactness and transferability, yielding a $+31.6\%$ N-ImageNet accuracy improvement.
For the attention-map alignment loss, we match the final-layer attention distributions of the two encoders using an MSE objective.
Combining all objectives slightly reduces transfer, suggesting that overly mixed signals may hinder convergence.

Overall, these results confirm that our multi-objective alignment design effectively improves semantic consistency between event and image modalities.

\begin{table}[t]
    \centering
    \caption{Ablation on alignment and training paradigm (frozen backbone).}
    \label{tab:ablation_align_mask_paradigm}
    \begingroup
    \setlength{\tabcolsep}{6pt}
    \renewcommand{\arraystretch}{1.05}
    \footnotesize
    \resizebox{\linewidth}{!}{%
    \begin{tabular}{lccc}
        \toprule
        Variant & Align & Parad. & DSEC (mIoU / mAcc) \\
        \midrule
        \rowcolor{gray!10}\textbf{Ours (2 stages)} & \Checkmark & AR  & 63.27 / \textbf{73.13} \\
        Event-only AR (Stage 2)          & \Checkmark    & AR   & \textbf{63.51} / 72.11 \\
        Ours - pretrain                 & \Checkmark    & None & 61.97 / 71.83 \\
        \rowcolor{gray!10}Ours - alignment              & \XSolidBrush & AR   & 52.38 / 60.97 \\
        Ours - alignment + finetune     & \XSolidBrush  & AR   & 58.24 / 66.82 \\
        \rowcolor{gray!10}Ours + aggregator~\cite{liu2025eventgpt} & \Checkmark & AR   & 61.24 / 71.53 \\
        Masked Modeling + alignment       & \Checkmark    & MAE  & 62.33 / 71.65 \\
        \bottomrule
    \end{tabular}%
    }
    \endgroup
    \vspace{-0.4cm}
\end{table}

\noindent \textbf{Training paradigm analysis.}
Removing image-to-event alignment means pretraining on the original DINOv2 feature sequence, which leads to a performance drop on DSEC from 63.27 to 52.38 mIoU, highlighting the importance of alignment.
Moreover, the two-stage design outperforms the single-stage variant by +1.30 mIoU, showing that autoregressive pretraining introduces temporal consistency beyond static alignment.
In the same setting of our framework, autoregressive (AR) pretraining surpasses masked modeling (MAE) by +0.94 mIoU and +1.48 mAcc, indicating stronger sequence-level reasoning.
Even without alignment (\textit{Ours - alignment + finetune}), our autoregressive variant still outperforms state-of-the-art approaches, yielding +5.72 mIoU over ECDP and +1.86 mIoU over ECDDP.

Furthermore, we ablate the downsampling of recent autoregressive video learners and multimodal LLMs, which often cluster spatio-temporal tokens for efficiency~\cite{ren2024arvideo, zhang2025videollama}.
Our controlled pretraining study (\textit{Ours + aggregator}) follows EventGPT~\cite{liu2025eventgpt}, which employs a spatio-temporal aggregator that performs independent pooling along the temporal and spatial dimensions and concatenates the results into a compact token sequence. 
The lower performance achieved by this aggregation demonstrates that aggressive token compression degrades dense prediction, which relies on high spatial fidelity.

%% file: sec/6_conclusion.tex
\section{Conclusion}
We presented GEP (Generative Event Pretraining), a unified framework that bridges large-scale VFMs and event-based representation learning.
By aligning event features with pretrained VFM representations and performing autoregressive sequence modeling on mixed event–image data, our method effectively transfers semantic knowledge from internet-scale image corpora to the event domain while preserving temporal sensitivity.

%% file: sec/5_ack.tex
\section*{Acknowledgment}
This work was supported by the European Union’s Horizon Europe Research and Innovation Programme under grant agreement No. 101120732 (AUTOASSESS) and the European Research Council (ERC) under grant agreement No. 864042 (AGILEFLIGHT).

%% file: sec/7_supplementary.tex
\clearpage
\begin{strip}
  \begin{center}
    {\LARGE Supplementary Material for}\\[6pt]
    {\LARGE \bfseries Generative Event Pretraining with Foundation Model Alignment}
  \end{center}
  \vspace{1em}
\end{strip}

\vspace{1em}
\section{Overview}
This supplementary document provides additional implementation details and
qualitative results that complement the main paper
``Generative Event Pretraining with Foundation Model Alignment''.
Specifically, Sec.~\ref{sec:method_details} presents extended method details, including training
hyperparameters and task-specific settings for semantic segmentation and depth
estimation, while Sec.~\ref{sec:qual_results} reports additional qualitative results for depth
prediction and attention visualization.

\section{Extended Method Details}
\label{sec:method_details}
\subsection{Alignment hyperparameters}
During the alignment stage, we fine-tune the pretrained backbone with a
lightweight optimization setup. We use the AdamW~\cite{loshchilov2017decoupled} optimizer with a batch size
of $32$ and an input resolution of $224 \times 224$. The base learning rate is
set to $3 \times 10^{-5}$ without weight decay, and we employ a linear warmup
for the first $100$ steps followed by a cosine decay schedule down to a
minimum learning rate of $0$. Training is performed for $2.5 \times 10^{5}$
steps.

To improve robustness, we apply moderate data augmentation, including polarity swap and
horizontal flipping, each with probability $0.5$. In addition, we use random
resized cropping with scale range $[0.25, 1.0]$ and a random upscale
operation (scale factor $2$) with probability $0.1$. All alignment
hyperparameters are summarized in Table~\ref{tab:alignment_hparams}.

\begin{table}
    \centering
    \caption{Hyperparameters for the alignment stage.}
    \label{tab:alignment_hparams}
    \begingroup
    \setlength{\tabcolsep}{4pt}
    \footnotesize
    \resizebox{0.8\linewidth}{!}{%
    \begin{tabular}{ll}
        \toprule
        Setting & Value \\
        \midrule
        \rowcolor{gray!10}
        optimizer             & AdamW \\
        Batch size            & 32 \\
        \rowcolor{gray!10}
        Learning rate         & $3 \times 10^{-5}$ \\
        Weight decay          & 0 \\
        Minimum learning rate & 0 \\
        \rowcolor{gray!10}
        Warmup steps          & 100 \\
        Training steps        & $2.5 \times 10^{5}$ \\
        \rowcolor{gray!10}
        Input resolution      & $224 \times 224$ \\
        Learning rate schedule& linear warmup \\
        \rowcolor{gray!10}
                              & cosine decay \\
        \midrule
        \multicolumn{2}{l}{\textbf{Data augmentation}} \\
        \midrule
        \rowcolor{gray!10}
        Polarity swap~\cite{yang2023event}                   & prob = 0.5 \\
        Horizontal flip                 & prob = 0.5 \\
        \rowcolor{gray!10}
        Random resized crop             & scale = [0.25, 1.00] \\
        Random upscale                  & prob = 0.1, scale = 2\\
        \bottomrule
    \end{tabular}%
    }
    \endgroup
\end{table}

\subsection{Pretraining hyperparameters}
For the pretraining stage, we adopt a similar optimization strategy but with
settings tailored to longer context and larger-scale training. We use AdamW
with a batch size of $8$ and a context window size of $4096$ (Following GPT2~\cite{radford2019language}). The base
learning rate is $5 \times 10^{-4}$ with weight decay of $1 \times 10^{-5}$.
We apply linear warmup for the first $100$ steps and then use a cosine decay
schedule towards an asymptote at $0$, training for
$2.5 \times 10^{5}$ steps in total. The full set of pretraining
hyperparameters is listed in Table~\ref{tab:pretrain_hparams}.

\begin{table}
    \centering
    \caption{Hyperparameters for the pretraining stage.}
    \label{tab:pretrain_hparams}
    \begingroup
    \setlength{\tabcolsep}{4pt}
    \footnotesize
    \resizebox{0.7\linewidth}{!}{%
    \begin{tabular}{ll}
        \toprule
        Setting & Value \\
        \midrule
        \rowcolor{gray!10}
        optimizer             & AdamW \\
        Batch size            & 8 \\
        \rowcolor{gray!10}
        Window size           & 4096 \\
        Learning rate         & $5 \times 10^{-4}$ \\
        \rowcolor{gray!10}
        Weight decay          & $1 \times 10^{-5}$ \\
        Minimum learning rate & 0 \\
        \rowcolor{gray!10}
        Warmup steps          & 100 \\
        Training steps        & $2.5 \times 10^{5}$ \\
        \rowcolor{gray!10}
        Learning rate schedule& linear warmup \\
                              & cosine decay \\
        \bottomrule
    \end{tabular}%
    }
    \endgroup
\end{table}

\subsection{Semantic Segmentation Settings}

For semantic segmentation, we adopt a lightweight decoding head that converts the sequence of visual tokens produced by the backbone into a full-resolution prediction map. The design consists of two main components: a linear patch-wise decoding stage and a shallow convolutional refinement stage. We use cross-entropy loss and dice loss~\cite{sudre2017generalised} with equal weights.

\paragraph{Linear Patch-wise Decoding.}
Vision Transformers (ViTs) operate on a grid of non-overlapping image patches. Given an input image with spatial resolution $H \times W$ and a patch size $P$, the encoder produces a sequence of
\begin{equation}
    L = \frac{H}{P} \times \frac{W}{P}
\end{equation}
tokens, each with feature dimension $D$. In contrast, dense prediction tasks provide supervision at the pixel level, i.e., one label per location in the original $H \times W$ grid.

To fully exploit this pixel-wise supervision, we adopt a simple \emph{linear patch-wise decoder}. For each token, we apply a shared linear projection that maps the $D$-dimensional feature into $C P^2$ channels, where $C$ is the number of classes. By reshaping and rearranging the outputs of all $L$ tokens, we recover a dense prediction map of shape $C \times H \times W$.

Formally, let $F \in \mathbb{R}^{L \times D}$ denote the token features from the ViT encoder. Our decoder learns a weight matrix
\begin{equation}
    W_{\text{dec}} \in \mathbb{R}^{D \times (C P^2)},
\end{equation}
and computes
\begin{equation}
    \hat{Y} = \mathrm{reshape}\big(F W_{\text{dec}}\big) \in \mathbb{R}^{C \times H \times W},
\end{equation}
where the reshape operation rearranges the per-token $C P^2$ outputs into their corresponding $P \times P$ spatial locations in the original image grid. This linear patch-wise decoding is computationally lightweight and allows us to leverage all pixel-level supervision without resorting to a heavy decoder.

\begin{figure*}
    \centering
    \includegraphics[width=0.98\textwidth]{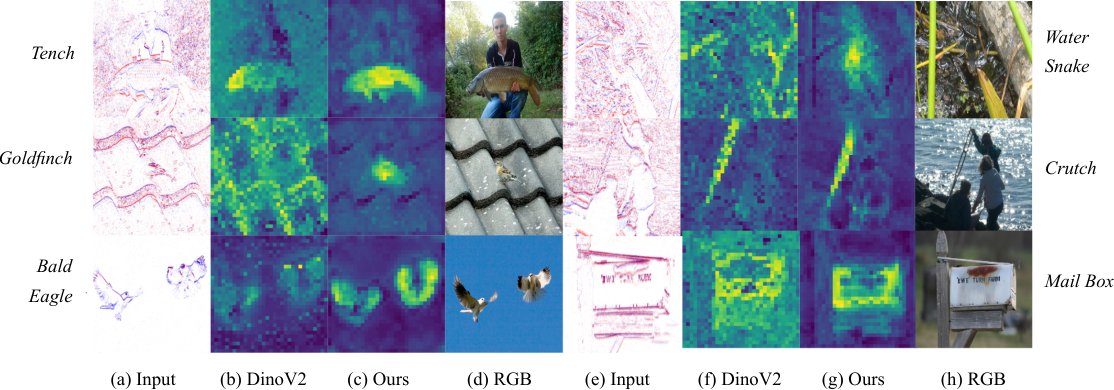}
    \caption{
    Visual comparison of attention maps on N-ImageNet~\cite{kim2021n} validation event inputs.
    We compare the attention responses of DINOv2~\cite{oquab2023dinov2} (b, f) and our method (c, g) given input events (a, e), alongside reference RGB images (d, h).
    Our approach demonstrates more precise localization of semantic structures compared to the diffuse activations of DINOv2.
    }
    \label{fig:vis_att_low}
\end{figure*}

\paragraph{Convolutional Refinement.}
To enhance spatial coherence and refine local details, the reconstructed feature map is further processed by a lightweight convolutional refinement module. This stage applies several $3{\times}3$ convolutions with nonlinear activations while preserving the original spatial resolution. The module introduces only 0.2\% (ViT-Small) and 0.07\% (ViT-Base) additional parameters, resulting in negligible impact on overall performance and computational cost, and is mainly employed to improve the visual quality of the predictions.

\subsection{Depth Estimation Settings}
Following prior work on monocular depth estimation~\cite{ranftl2021vision,yang2024event}, we supervise training with the scale-invariant loss of~\cite{eigen2014depth} and the multi-scale scale-invariant gradient matching loss of~\cite{hidalgo2020learning}, using fixed loss weights of 1 and 0.25, respectively. The network is optimized to predict normalized log-depth values during training, and these predictions are converted back to metric depth at inference time.

\paragraph{Decoder backbone.}
Given a batch of encoder tokens
$T \in \mathbb{R}^{B \times L \times C}$,
we treat $B$ as the batch size, $L$ as the number of tokens per image, and
$C$ as the channel dimension of each token.
These tokens lie on a patch grid of size $H_p \times W_p$, obtained from an
input image of spatial resolution $H \times W$ using a patch size $p$, so that
$H_p = H / p$ and $W_p = W / p$.

We first apply a linear projection to each token and reshape the result into
a low-resolution feature map
$F_0 \in \mathbb{R}^{B \times C_0 \times H_p \times W_p}$,
where $C_0$ is the number of feature channels used in the decoder.
On top of $F_0$, we build a convolutional decoder with $S$ stages,
indexed by $s = 1,\dots,S$.
Each stage is implemented as a residual-like block, denoted
$\mathrm{DecBlock}_s$, that consists of two $3{\times}3$ convolutions with
GELU activations and an identity skip connection.
The input to stage $s$ is $F_{s-1}$ and the output is $F_s$.

Each decoder stage has a target spatial resolution $(H_s, W_s)$.
These resolutions gradually increase from the patch resolution
$(H_p, W_p)$ at the first stage up to the full image resolution
$(H, W)$ at the last stage.
If the output $F_s$ of stage $s$ does not match $(H_s, W_s)$, we use a
bilinear interpolation operator, denoted $\mathrm{Interp}(\cdot; H_s, W_s)$,
to upsample it to the target size and obtain a feature map
$\tilde F_s$ at resolution $(H_s, W_s)$.

To aggregate information across scales, all stage outputs are first brought to
full resolution $(H, W)$ using the same interpolation operator and then
concatenated along the channel dimension.
This yields a fused feature map
$F_{\mathrm{fuse}} \in \mathbb{R}^{B \times C_{\mathrm{fuse}} \times H \times W}$,
where $C_{\mathrm{fuse}}$ is the total number of channels after concatenation.

Finally, a small convolutional prediction head is applied to
$F_{\mathrm{fuse}}$.
This head consists of two $3{\times}3$ convolutions with GELU activations,
followed by a $1{\times}1$ convolution and a sigmoid function.
The output is a normalized log-depth map
$Y_{\mathrm{norm}} \in [0,1]^{B \times 1 \times H \times W}$,
where the single channel corresponds to normalized log-depth for each pixel.

\paragraph{Log-depth parameterization.}
We convert the normalized log-depth $Y_{\mathrm{norm}}$ into metric depth using
a fixed minimum and maximum depth range.
Let $d_{\min} > 0$ and $d_{\max} > d_{\min}$ denote, respectively, the minimum
and maximum metric depth considered during training.
We define their logarithms
$\ell_{\min} = \log d_{\min}$ and $\ell_{\max} = \log d_{\max}$,
and the log-range
$\Delta \ell = \ell_{\max} - \ell_{\min}$.
For each pixel, the normalized log-depth prediction $Y_{\mathrm{norm}}$
is first mapped to the corresponding log-depth value
$\ell = Y_{\mathrm{norm}} \,\Delta \ell + \ell_{\min}$,
and the final metric depth prediction is obtained as
\begin{equation}
D = \exp(\ell)
\in \mathbb{R}^{B \times 1 \times H \times W}
\end{equation}
Thus, $D$ contains the predicted depth (in meters) at each pixel.
This log-depth parameterization improves numerical stability over large
depth ranges and makes the predictions more robust to global scale shifts,
since uniform shifts in log-depth correspond to multiplicative changes in
metric depth.

\textbf{Multi-scale log-depth gradient loss.}
To further regularize the spatial structure of the prediction, we match
log-depth gradients at multiple resolutions.
We consider a set of scales $\mathcal{S}$ (e.g., different downsampling
factors).
For each scale $s \in \mathcal{S}$, we downsample the predicted depth,
the ground-truth depth, and the validity mask with a fixed operator
$\mathrm{Down}_s(\cdot)$:
\begin{equation}
D^{(s)} = \mathrm{Down}_s(D),
\end{equation}
\begin{equation}
(D^\star)^{(s)} = \mathrm{Down}_s(D^\star),
\end{equation}
\begin{equation}
M^{(s)} = \mathbf{1}\{\mathrm{Down}_s(M) > 0.5\}.
\end{equation}

\paragraph{Training loss.}
Supervision is provided by LiDAR depth measurements
$D^\star \in \mathbb{R}^{B \times 1 \times H \times W}$ and a binary validity
mask $M \in \{0,1\}^{B \times 1 \times H \times W}$, where $M_i = 1$ indicates
that a valid ground-truth depth value is available at pixel $i$ and
$M_i = 0$ otherwise.
All losses are computed only on pixels with $M_i = 1$.

To concisely express masked averages, we define a masked mean operator.
Given any quantity $X$ defined per pixel (and per batch element), and a
corresponding binary mask $M$, we write
\begin{equation}
\label{eq:masked-mean}
\langle X \rangle_M
= \frac{\sum_{i} M_i X_i}{\sum_{i} M_i + \varepsilon},
\end{equation}
where $i$ indexes all spatial positions and batch elements, and
$\varepsilon$ is a small constant used to avoid division by zero when
no valid pixels are present.

\textbf{Scale-invariant log RMSE.}
Following Eigen et al~\cite{eigen2014depth}., we measure the discrepancy between predicted and
ground-truth depths in log space.
For each valid pixel $i$, we define the log-depth residual
\[
y_i = \log D_i - \log D^\star_i,
\]
where $D_i$ and $D^\star_i$ denote the predicted and ground-truth depth,
respectively.
Using the masked mean operator in Eq.~\ref{eq:masked-mean}, the
scale-invariant log-depth loss is
\begin{equation}
L_{\mathrm{silog}}
= \sqrt{
  \big\langle y_i^2 \big\rangle_M
  - \lambda \,\big\langle y_i \big\rangle_M^2
}.
\end{equation}
Here, $\lambda \in [0,1]$ is a scalar hyperparameter that controls the
degree of scale invariance; in our experiments we use $\lambda = 0.85$.
The first term encourages small squared log-errors, while the second term
subtracts a global bias term, making the loss invariant (up to $\lambda$)
to additive shifts in log-depth, i.e., multiplicative changes in depth scale.

At scale $s$, we compute horizontal and vertical log-depth gradient
residuals using finite differences:
\begin{equation}
G_x^{(s)} = \nabla_x \log D^{(s)} - \nabla_x \log (D^\star)^{(s)},
\end{equation}
\begin{equation}
G_y^{(s)} = \nabla_y \log D^{(s)} - \nabla_y \log (D^\star)^{(s)}.
\end{equation}
The corresponding masked mean absolute errors are
\begin{equation}
E_x^{(s)} = \big\langle |G_x^{(s)}| \big\rangle_{M^{(s)}},
\end{equation}
\begin{equation}
E_y^{(s)} = \big\langle |G_y^{(s)}| \big\rangle_{M^{(s)}},
\end{equation}
where $\langle \cdot \rangle_{M^{(s)}}$ denotes the masked mean defined
analogously to Eq.~\eqref{eq:masked-mean}, but using the mask $M^{(s)}$.

The gradient loss at scale $s$ and the final multi-scale gradient loss are
\begin{equation}
L_{\mathrm{grad}}^{(s)} = E_x^{(s)} + E_y^{(s)},
\end{equation}
\begin{equation}
L_{\mathrm{ms\_grad}}
= \frac{1}{|\mathcal{S}|} \sum_{s \in \mathcal{S}} L_{\mathrm{grad}}^{(s)},
\end{equation}
where $|\mathcal{S}|$ is the number of scales in $\mathcal{S}$.

\paragraph{Final objective.}
The total depth training objective is a weighted sum of the scale-invariant
log RMSE and the multi-scale gradient loss:
\begin{equation}
L_{\mathrm{total}}
= w_{\mathrm{silog}} \, L_{\mathrm{silog}}
+ w_{\mathrm{ms\_grad}} \, L_{\mathrm{ms\_grad}},
\end{equation}
where $w_{\mathrm{silog}}$ and $w_{\mathrm{ms\_grad}}$ are non-negative
scalar hyperparameters that balance the two terms.
Gradients are backpropagated only through pixels with $M_i = 1$ (and their
downsampled counterparts), so invalid or missing depth values do not
influence learning.

\section{Computational Efficiency and Temporal Analysis}
\label{sec:efficiency_temporal}

\subsection{Computational Efficiency}
As quantified in \cref{tab:pretrain_cost,tab:finetune_perf}, GEP achieves 10$\times$ and $\sim$65$\times$ efficiency gains in pretraining and finetuning respectively with superior performance. The GPU hours are computed with a single RTX8000.

\begin{table}[h]
    \centering
    \caption{\textbf{Pre-training Efficiency \& Data.} Comparison of backbone architectures, pre-training paradigm, resolution, compute cost, data scale, and peak memory usage.}
    \label{tab:pretrain_cost}
    \resizebox{\linewidth}{!}{
    \begin{tabular}{l c c c c c c}
        \toprule
        \multirow{2}{*}{\textbf{Method}} & \textbf{Back-} & \textbf{Pretrain} & \textbf{Pretrain} & \textbf{Pretrain} & \textbf{Pretrain} & \textbf{Peak} \\
         & \textbf{bone} & \textbf{Mode} & \textbf{Res.} & \textit{EFLOPs (h)} & \textit{Data Size} & \textbf{GPU Mem.} \\
        \midrule
        TESPEC~\cite{mohammadi2025tespec} & Swin-T & Unsup. & $360{\times}640$ & 31.7 (540h) & $\sim$3.5 TB & - \\
        STP~\cite{liang2025efficient} & Swin-T & Sup. & $224{\times}224$ & - & \textbf{0.57 TB} & - \\
        \textbf{GEP (Ours)} & ViT-S & Unsup. & $224{\times}224$ & \textbf{3.2 (54h)} & 0.94 TB & \textbf{17G} \\
        \bottomrule
    \end{tabular}
    }
\end{table}

\begin{table}[h]
    \centering
    \caption{Segmentation Performance \& Cost.}
    \label{tab:finetune_perf}
    \resizebox{\linewidth}{!}{
    \begin{tabular}{l c c c c c}
        \toprule
        \multirow{2}{*}{\textbf{Method}} & \textbf{Head} & \textbf{Trainable Param.} & \textbf{Finetune (h)} & \multicolumn{2}{c}{\textbf{DSEC Seg.}} \\
         &  &  &  & \textbf{mIoU} & \textbf{mAcc} \\
        \midrule
        TESPEC~\cite{mohammadi2025tespec} & UperNet & 45M & 2.7 (46h) & 62.77 & 70.61 \\
        STP~\cite{liang2025efficient} & UperNet & 42M & - & 62.05 & - \\
        \textbf{GEP (ours)} & Linear & \textbf{22M} & \textbf{0.04 (0.7h)} & \textbf{65.22} & \textbf{74.66} \\
        \bottomrule
    \end{tabular}
    }
\end{table}

\subsection{Temporal Resolution Analysis}
Our approach unlocks VFM semantics \textit{without} compromising temporal fidelity. As empirically proven in \cref{tab:temporal_ablation}, our model supports \textit{flexible time sampling}: it generalizes robustly to unseen high-temporal resolutions ($N=10$ bins), demonstrating the capability to flexibly adapt the integration window $\Delta t$ to extreme motion speeds on the fly. Furthermore, it is essentially representation-agnostic and can seamlessly adopt inputs like Voxel Grids.

\begin{table}[h]
    \centering
    \caption{\textbf{Temporal Resolution Ablation.} All experiments are transferred based on 10\% schedule pretrained pilot runs.}
    \label{tab:temporal_ablation}
    \resizebox{\linewidth}{!}{
    \begin{tabular}{l c c c}
        \toprule
        \textbf{Pre-training Strategy} & \textbf{Eval. Resolution} & \multicolumn{2}{c}{\textbf{DSEC Seg.}} \\
        \textit{(Bins per RGB interval)} & \textit{(Bins per RGB interval)} & \textbf{mIoU} & \textbf{mAcc} \\
        \midrule
        Fixed ($N=1$) (Ours) & $N=1$ (Ours) & 42.98 & 53.09 \\
        \midrule
        \multirow{3}{*}{Dynamic ($N \in \{1, 3, 5\}$)} & $N=1$ (Ours)& 42.67 & 52.02 \\
         & $N=10$ (Unseen) & 42.25 & 51.39 \\
         & $N=10$ (Random) & 21.31 & 28.58 \\
        \bottomrule
    \end{tabular}
    }
\end{table}

\section{More Qualitative Results}
\label{sec:qual_results}
\subsection{Depth Estimation}
\begin{figure}
  \centering
  \includegraphics[width=0.47\textwidth]{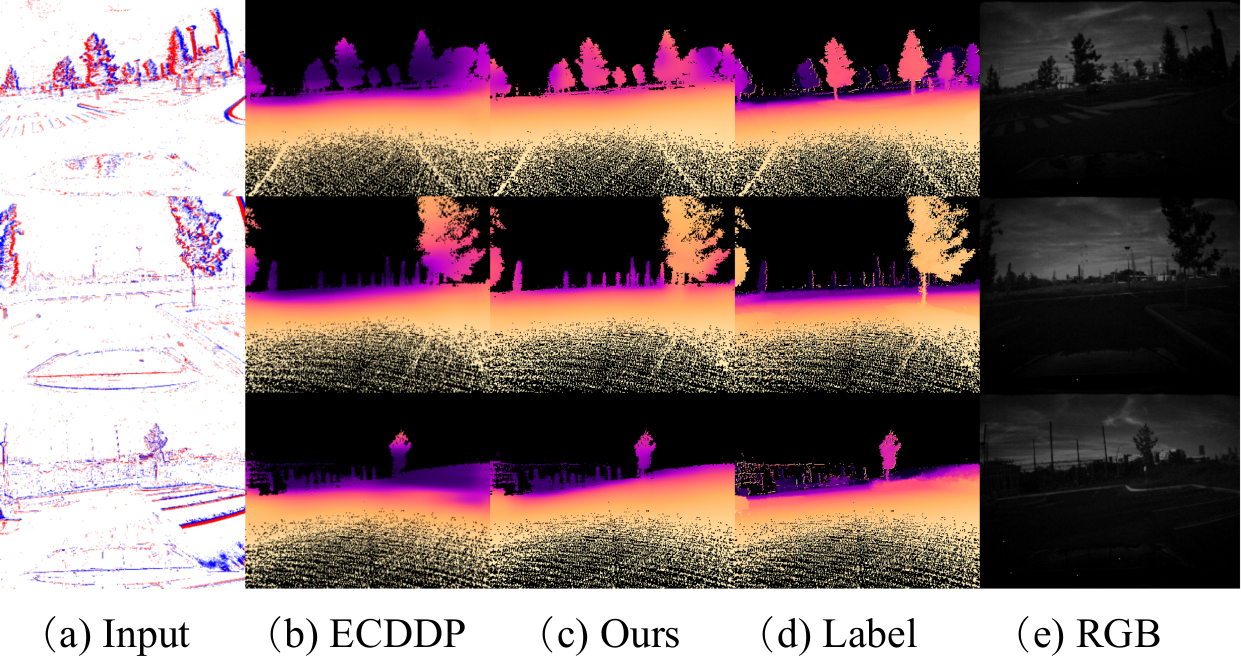}
  \caption{
  Qualitative comparison of depth estimation on MVSEC~\cite{zhu2018multivehicle}.
  From left to right: (a) event-based input, (b) ECDDP baseline,
  (c) our method, (d) ground-truth depth, and (e) corresponding reference RGB
  images (not used). Our model produces depth maps that more closely match the
  ground truth, with sharper object boundaries and fewer artifacts
  on thin structures such as trees and poles.
  }
  \label{fig:depth_vis}
\end{figure}

Figure~\ref{fig:depth_vis} presents qualitative depth estimation
results on several outdoor driving scenes.
Compared to the ECDDP baseline, our method recovers a smoother road
surfaces and sharper discontinuities at object boundaries, while
preserving thin structures such as trees and poles.
The predicted depth maps align better with the LiDAR ground truth,
illustrating the benefit of our representation for event-based depth
estimation.
\subsection{Attention Map}

In addition to depth estimation, we further analyze the discriminative power of our learned features by visualizing attention maps in Figure~\ref{fig:vis_att_low}.
For each sample, we show the input event frame, the attention responses produced by DINOv2~\cite{oquab2023dinov2} (our teacher VFM during alignment) and our method, and the corresponding RGB image.
As observed in the figure, our approach focuses more accurately on semantically meaningful structures such as object contours, fine edges, and distinctive regions.
In contrast, DINOv2 exhibits more diffuse or noisy activations and often fails to localize these structures precisely.
These qualitative results indicate that our representation extracts more discriminative features from event data and aligns better with the underlying scene semantics.

